\newif\iffinal
\finaltrue

\documentclass[final,1p,times,onecolumn]{elsarticle}
\usepackage{amssymb}
\usepackage{amsmath}
\usepackage{amstext}
\usepackage{amsopn}
\usepackage{ifpdf}
\usepackage{boxedminipage}
\usepackage{xcolor}
\usepackage{lineno}
\usepackage{multirow}
\usepackage{version}
\usepackage{array}
\usepackage{url}
\usepackage{amsmath}
\usepackage{graphics}
\usepackage{epsfig}
\usepackage{mathtools}
\usepackage{tabularx}

\usepackage{listings}
\usepackage{color}
\usepackage{import}

\ifpdf
\usepackage[%
  pdfsubject={Preprint submitted to Elsevier},%
  pdfstartview=FitH,%
  bookmarks=true,%
  bookmarksopen=true,%
  bookmarksnumbered=true,%
  bookmarksopenlevel=4,%
  breaklinks=true,%
  colorlinks=true,%
  linkcolor=blue,%
  anchorcolor=blue,%
  citecolor=blue,%
  filecolor=blue,%
  menucolor=blue,%
  pagecolor=blue,%
  urlcolor=blue]{hyperref}
\else
\usepackage[%
  breaklinks=true,%
  colorlinks=true,%
  linkcolor=blue,%
  anchorcolor=blue,%
  citecolor=blue,%
  filecolor=blue,%
  menucolor=blue,%
  pagecolor=blue,%
  urlcolor=blue]{hyperref}
\fi

\usepackage[ruled,vlined]{algorithm2e}

\newcounter{algorithm}
\begin{document}

\begin{frontmatter}

  \title{ Explainable AI (XAI) in Image Segmentation in Medicine, Industry, and Beyond: A Survey}

\author{Rokas Gipi\v{s}kis$^1$}
\ead{rokas.gipiskis@mif.vu.lt}\cortext[cor]{Corresponding author.}

\author{Chun-Wei Tsai$^2$}
\author{Olga Kurasova$^1$}
\address{$^1$Vilnius University, Institute of Data Science and Digital Technologies}
\address{$^2$National Sun Yat-sen University, Department of Computer Science and Engineering}


  \begin{abstract} 
      Artificial Intelligence (XAI) has found numerous applications in
      computer vision. While image classification-based explainability
      techniques have garnered significant attention, their
      counterparts in semantic segmentation have been relatively
      neglected. Given the prevalent use of image segmentation,
      ranging from medical to industrial deployments, these techniques
      warrant a systematic look. In this paper, we present the first
      comprehensive survey on XAI in semantic image segmentation. This
      work focuses on techniques that were either specifically
      introduced for dense prediction tasks or were extended for them
      by modifying existing methods in classification. We analyze and
      categorize the literature based on application categories and
      domains, as well as the evaluation metrics and datasets used. We
      also propose a taxonomy for interpretable semantic segmentation,
      and discuss potential challenges and future research
      directions.

  \end{abstract}

\end{frontmatter}

\section{Introduction}

In the past decade, Artificial Intelligence (AI) systems have achieved impressive results, most notably in natural language processing and computer vision. The performance of such systems is typically measured by evaluation metrics that vary depending on the task but aim to assess the system's outputs. Today's leading AI systems largely rely on deep learning (DL) models, multi-layered neural networks that tend to exhibit increasingly complicated structures in terms of model parameters. The growing complexity of such systems resulted in them being labeled as ``black boxes.'' This highlights that the evaluation metric does not show the full picture: even if its measurement is correct, it does not give insights into the inner workings of the model.  

The field of explainable AI (XAI) encompasses different branches of methods that attempt to give insights into a model's inner workings, explain outputs, or make the entire system more interpretable to end users, such as human decision-makers. There is ongoing debate regarding XAI terminology. Concepts like interpretability, explainability, understanding, reasoning, and trustworthiness are challenging to formalize. While some authors use ``interpretable'' and ``explainable'' interchangeably \cite{miller2019explanation}, others distinguish between the two \cite{rudin2019stop}, \cite{roscher2020explainable}. When the distinction is made, it is usually to demarcate post-hoc explanations, a type of XAI techniques applied to the already-trained model, and inherently interpretable models \cite{rudin2019stop}. This way interpretability becomes associated with the transparency of the model itself and depends on the ease with which one can interpret the model. For instance, a simple decision tree-based model might be considered more interpretable than a DL model composed of millions of parameters, provided that the former is not too deep. Explainability, in contrast, is often limited to understanding the model's results rather than the model as a whole. While we acknowledge such distinction, throughout this survey we will use ``interpretable'' and ``explainable'' synonymously, reserving more specific ``architecture-based'' and ``inherently interpretable'' terms when discussing model-specific XAI modifications. This is because not many of the surveyed papers use the term interpretability in a second sense. Since most papers in explainable segmentation do not make this distinction, this might avoid unnecessary confusion when discussing their contents. It should also be noted that interpretability and ease of understanding vary according to the specific audience, whether it be the general public or a more specialized group with specific training, such as radiologists.

\begin{figure}[htb]
    \centering
    \includegraphics[width=0.8\textwidth]{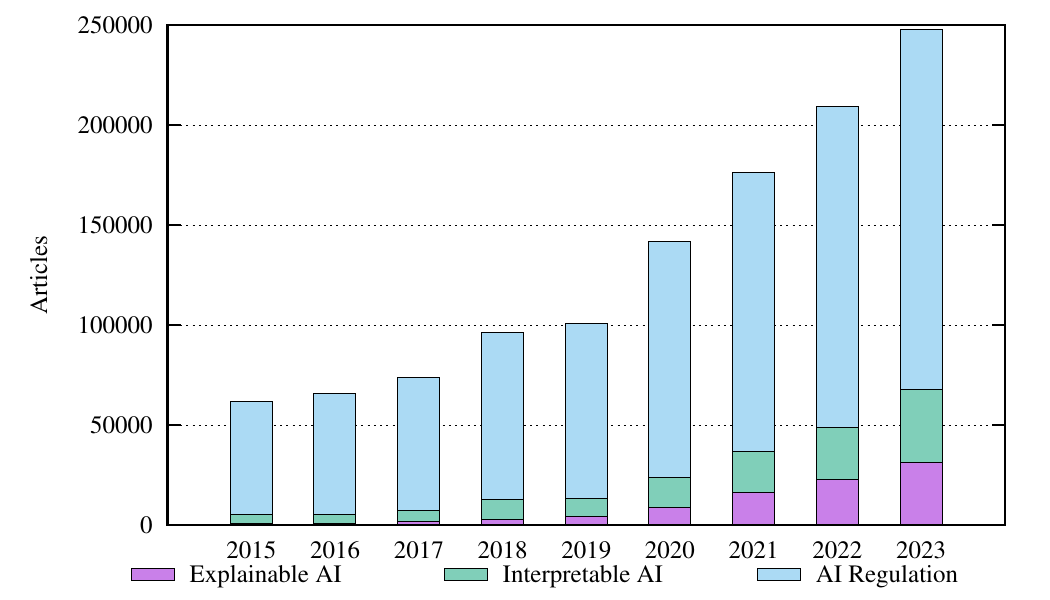}
 
    \caption{Publications with ``explainable AI,'' ``interpretable AI,'' and ``AI regulation'' as keywords. Publication data gathered from app.dimensions.ai}
    \label{fig:publications}
\end{figure}

XAI is not a new development, particularly in rule-based expert systems \cite{shortliffe1975model}, \cite{swartout1991explanations} and machine learning (ML) \cite{friedman2001greedy}, but it has experienced unprecedented growth ever since the revived interest \cite{krizhevsky2012imagenet} in neural networks. This growth correlates with the increasing interest in DL and is further driven by 1)~the need for trustworthy models due to widely expanding industrial deployments; 2)~bureaucratic and top-down political emphasis on AI regulation; and 3)~concerns within the ML safety community \cite{amodei2016concrete} about the general trajectory of AI development in the short and long runs. AI deployment is increasing across different sectors, and is significant both in terms of its size and impact. 
According to AI Index Report 2023 \cite{maslej2023artificial}, the proportion of companies adopting AI has more than doubled from 2017 to 2022. In 2022, the medical and healthcare sectors have attracted the most investment, with a total of 6.1 billion dollars \cite{maslej2023artificial}. IBM Global AI Adoption Index
2023 \cite{ibm2023report}, conducted by Morning Consult on behalf of IBM,
%
indicates that about 42\% of their surveyed ($>1,000$ employees) enterprise-scale companies reported actively deploying AI, and an additional 40\% exploring and experimenting with AI, out of which 59\% reported an acceleration in their rollout or investments.  Even with rapid deployment, critical high-impact sectors have to move at a slower pace. One could expect even more healthcare-related applications and clinical deployments if AI methods were more interpretable. To a large extent, this applies to other industries as well. According to the same IBM report, most of the surveyed IT professionals (83\% among companies already exploring or deploying AI) stated that it is important to their business to explain how their AI reached the decision. Another accelerating trend is that of AI regulation (Fig. \ref{fig:publications}). The recent survey \cite{curtis2023ai} indicates that 81\% of respondents (\textit{N} $>$ 6,000) expect some form of external AI regulation, with 57-66\% of respondents reporting that they would be more willing to use AI systems if trustworthiness-assuring mechanisms were in place. AI trustworthiness and transparency are further emphasized in regulatory discussions, ranging from the EU's AI Act \cite{commisionregulation} to AI executive order \cite{biden2023executive} in the United States.

 XAI in image segmentation is a relatively new field, with the first articles on the subject appearing in the late 2010s \cite{wickstrom2020uncertainty}, \cite{hoyer2019grid}, \cite{vinogradova2020towards}. Since then, the topic has gained more attention. Semantic image segmentation is an essential task in computer vision, with applications ranging from autonomous driving \cite{feng2020deep} to medical image analysis \cite{asgari2021deep}. Its study is further motivated by the rapidly growing remote sensing and video data. Increasing deployments in medical AI are also contributing to the need for explainable segmentation. Both radiologists and surgeons need to know accurate boundaries for the anatomical structures of interest. Precise and reliable segmentation is required when working with most pathologies in different imaging modalities, ranging from magnetic resonance imaging (MRI) to computed tomography (CT).  

Segmentation is commonly viewed as a dense prediction task where classification is performed on a pixel level. However, most XAI literature so far has focused on image classification tasks. Nonetheless, a growing number of works address the issue of interpreting semantic segmentation results by either extending classification-based methods or by proposing their own modifications. Two Ph.D. dissertations \cite{vinogradova2023explainable}, \cite{mullan2023deep} on XAI in image segmentation have been written in the past year. Image segmentation methods have been reviewed in the medical domain \cite{hasany2024post}, however, the focus has been just on the post-hoc techniques. In this work, we provide the first comprehensive survey in the area of explainable semantic segmentation, encompassing different application domains and all XAI method groups. Throughout the paper, when referring to the date of the publication, we indicate the date of its first online appearance, including the preprint version where applicable.

This survey offers the following contributions:

\begin{itemize}
   
  \item  A comprehensive review of up-to-date publications, covering both their theoretical contributions and practical applications. 
  \item A taxonomy to distinguish various interpretability techniques based on five subgroups.
  \item Analysis of the literature based on evaluation metrics, datasets used, and application domains.
  \item A detailed discussion of open issues and identification of future research directions.
\end{itemize}

The paper is structured as follows. Section 2 begins with the general scope of the problem and then provides the background for the fields of XAI and semantic image segmentation. Section 3 reviews the most important taxonomical dichotomies in classification and introduces a method-centered taxonomy for XAI in image segmentation. Section 4 presents illustrative examples of each method group, includes formalizations for gradient-based and perturbation-based methods, and outlines XAI evaluation metrics.  Short summaries of each method with their main contributions, grouped by application area, are provided in Section 5. Lastly, Section 6 points out future research directions, while Section 7 draws the main conclusions from this study.

\section{Background}
\subsection{Development of the Field of XAI in Computer Vision}

 There is a great variety of XAI methods in classification, with new techniques being proposed weekly. Typically, these methods employ some form of feature attribution, indicating the model's sensitivity or insensitivity to various features, such as certain pixel configurations in the input space. The most popular explainable classification methods, still influential in today's DL models, fall into gradient-based or perturbation-based categories. Here, we only highlight key developments, particularly focusing on the methods that have influenced interpretable image segmentation. For an accessible introduction to and treatment of explainable classification, we refer the reader to \cite{molnar2020interpretable}. For a more detailed survey on these topics, \cite{zhang2021survey} is also recommended.

 The first gradient-based explainability techniques for classification in convolutional neural networks (CNNs) are proposed in \cite{simonyan2013deep}. The initial method generates artificial images that maximize the score for the selected class of interest. The second method, also referred to as vanilla gradient, produces a saliency map that highlights important regions in the input space. This is based on the gradient for the class of interest with respect to the input. The authors also observe that this method can be used for weakly supervised segmentation. This marks the possibility to use XAI tools instrumentally, not just for the sake of explainability. In \cite{selvaraju2017grad}, the influential Grad-CAM technique is introduced. Its calculation is based on the gradient flow into the last convolutional layer. Since Grad-CAM is calculated for intermediate model activations, the resulting explanation needs to be upsampled. This upsampling process might negatively impact the quality of pixel-level explanations \cite{koker2021u}. Similar to \cite{simonyan2013deep}, the Grad-CAM technique also demonstrates the potential for instrumental use in weakly supervised localization.

Another area of explainable classification methods encompasses occlusion or perturbation-based techniques, such as occlusion sensitivity measurements \cite{zeiler2014visualizing}, LIME \cite{ribeiro2016should}, SHAP \cite{lundberg2017unified}, and RISE \cite{petsiuk2018rise}. In \cite{zeiler2014visualizing}, occlusion sensitivity is introduced. It proposes systematically occluding the input image with a smaller grey filter and measuring the effect on the model's output. The likelihood of the model classifying the image as belonging to the actual class should decrease when the object of that class is occluded in the input space. Other noteworthy methods in explainable classification have focused on optimization. Activation maximization, previously proposed in \cite{erhan2009visualizing}, initially focused on Restricted Boltzmann Machines, a type of unsupervised models. In \cite{simonyan2013deep}, it has been specifically implemented in supervised classification models. In \cite{olah2017feature}, this technique was further popularized by demonstrating the results across different network layers. Unlike the previously discussed XAI techniques, this type of explanation method can be described as global because the generated image does not depend on a particular input image but rather on the model's internal weights.

\subsection{Specifics of Semantic Segmentation}

Most of the literature on interpretable computer vision focuses on classification. However, DL-based semantic segmentation techniques have achieved significant results. Classical encoder-decoder models such as U-Net \cite{ronneberger2015u} or SegNet \cite{badrinarayanan2017segnet} as well as their modifications, have been deployed in various fields. Vision transformer-based segmentation architectures have also been proposed \cite{strudel2021segmenter}. There have even been attempts to combine these two approaches \cite{chen2021transunet}. During semantic segmentation, class labels are assigned to each pixel, and the output is typically the same resolution as the input image. Modern segmentation models can be composed of millions of parameters, making their interpretation difficult and often resulting in their description as ``black boxes.''

Interpretability in semantic image segmentation is a challenging area of study. On one hand, it can be viewed as an extension of a relatively intuitive interpretable classification. However, it requires combining the relative influence of each classified pixel of interest. On the other hand, interpreting its own explanations is not so straightforward or intuitive. One problem with interpretability methods, not limited to semantic segmentation, is that we generally lack ground truths for the explanations. Furthermore, it is uncertain what the ideal explanation should look like or whether one interpretable instance can be limited to a single explanation. However, in the case of classification, we can at least have some candidates for good explanations. Based on this, qualitative human-based studies \cite{kim2022hive} can be conducted. Conducting a similar study for semantic segmentation is more complex, as it is less clear what constitutes good explanation candidates: should the interpretability saliency map focus on the entire area of the class of interest or just its boundaries? Can there be instances where the most salient features are outside the class area? What if the segmentation area is correct, but the attributed class is not? Moreover, semantic image segmentation is notorious for inter-observer variability, especially in manual delineations in medical images. One way to demonstrate the usefulness of explainable segmentation is to detect instances where the segmentation of one semantic class appears heavily dependent on the presence of different class pixels, whether nearby or otherwise. In \cite{vinogradova2020towards}, such a case is demonstrated when the U-Net detects the sky primarily due to the nearby trees, which belong to the 'Nature' class. Interpretable semantic segmentation techniques prove most useful when the segmentation is incorrect.

\begin{figure*}[!ht]
    \centering
    \footnotesize
     \begin{tabular}{cc}
        \includegraphics[width=0.3\textwidth]{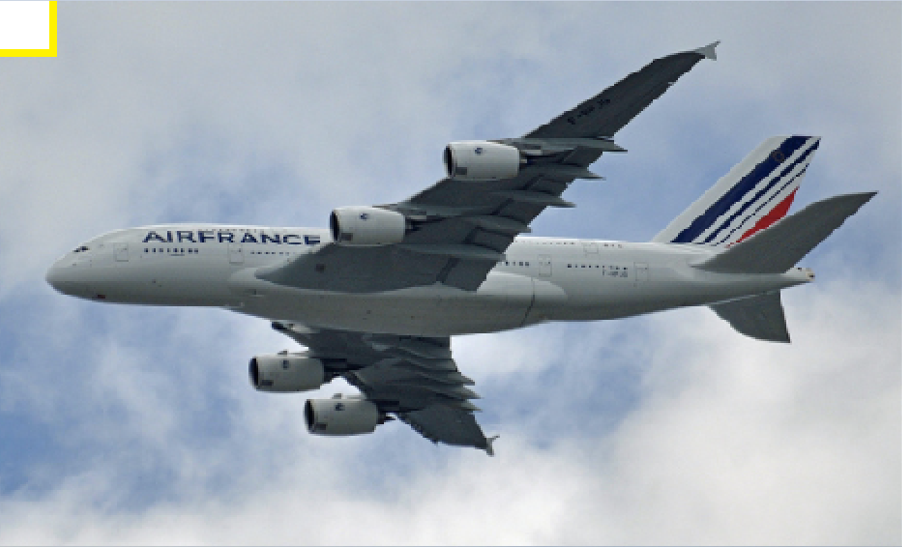} & \includegraphics[width=0.33\textwidth]{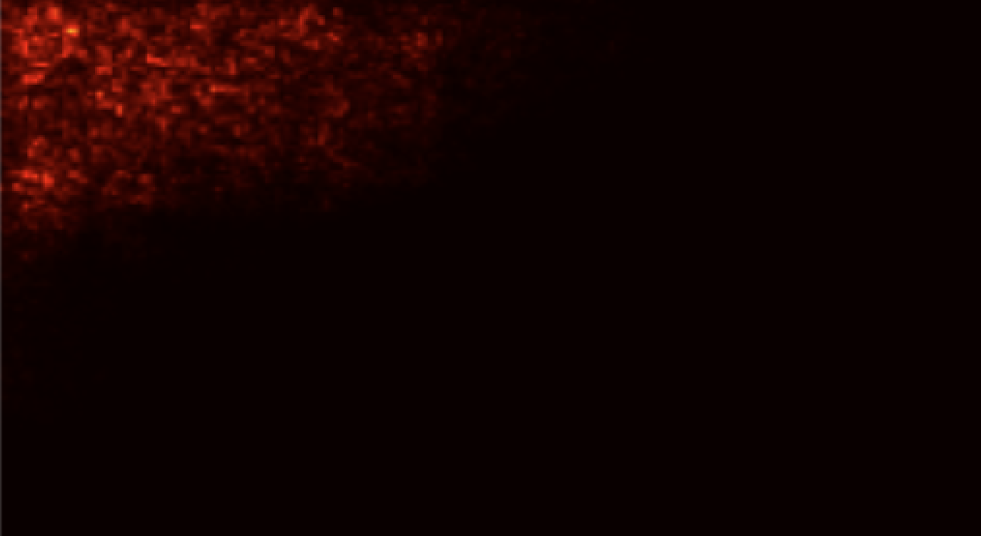}\\
        (a) & (b)\\
        \includegraphics[width=0.3\textwidth]{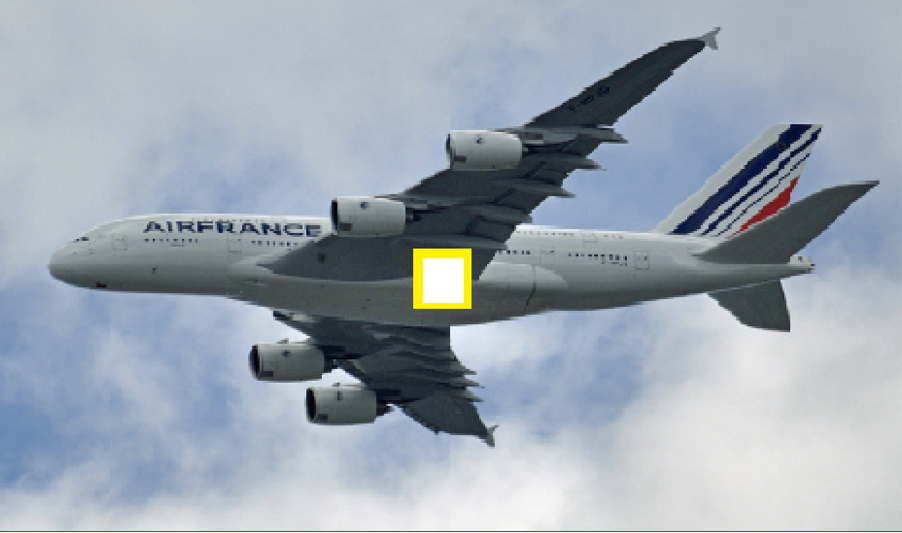} & \includegraphics[width=0.33\textwidth]{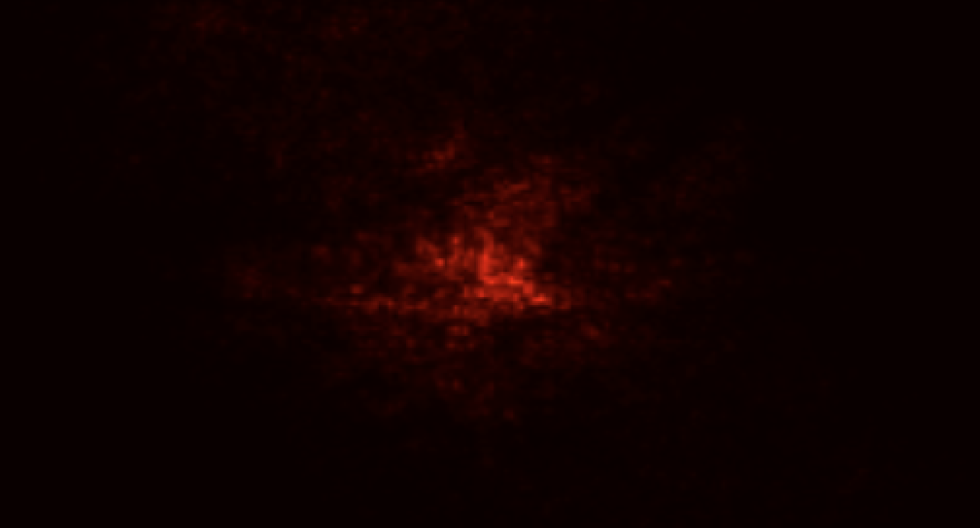}\\
        (c) & (d)\\
    \end{tabular} 
    \caption{ Explanation for single pixels: the selected pixels (top leftmost and centermost) are shown on the left, with their corresponding gradient-based explanations on the right.}
    \label{fig:pixel}
\end{figure*}

Since we can frame the segmentation task in terms of classification, it is relatively easy to apply explainable classification methods to it, focusing on a single pixel as seen in Fig. \ref{fig:pixel}. For instance, a gradient for the selected output pixel of a chosen class can be calculated with respect to the entire input image. However, an explanation map for the classification of a single pixel is not particularly useful. It is less accessible to the human interpreter, as evaluating thousands of different explanations for just a single class in a single image would be required. Therefore, we need to consider the effects of a larger number of pixels. Most popular explainable segmentation techniques operate under the underlying assumption of pixel importance. This assumption is particularly relevant to perturbation-based methods, where introducing noise to important pixels would degrade a model's performance more significantly than adding it to less critical pixels. To explain the whole image (i.e., all pixels) instead of just a single pixel, most explainable segmentation techniques must visualize the relative contributions of all pixels simultaneously. Otherwise, the analysis of separate single-pixel-based explanation maps would be too tedious. The most popular way to do it involves using logit values, unnormalized probabilities before the Softmax layer, typically used in classification. This could be achieved by summing up the logits of the class of interest for the pixels of interest, for instance. This new scalar value can then be used when generating a single explanation for the entire image, just like in the case of a single pixel.

\subsection{Limitations}
Feature attribution and saliency-based XAI methods in particular have faced criticism \cite{adebayo2018sanity}, \cite{kindermans2019reliability}, \cite{saporta2022benchmarking}. Although these criticisms have solely focused on explainable classification, they deserve a thorough examination as they could also extend to segmentation. Some of the XAI methods act as regular edge detectors, independently from the underlying model and training dataset. This independence is troubling because a local post-hoc XAI method should explain a specific model's prediction for a particular data point. In \cite{bilodeau2022impossibility}, limitations of feature attribution methods such as SHAP and integrated gradients are emphasized both theoretically and empirically, showing that they cannot reliably infer counterfactual model behavior. The authors observe that the analyzed attribution methods resemble random guessing in tasks like algorithmic recourse and spurious feature identification. Similar experimental results are observed with gradients, SmoothGrad \cite{smilkov2017smoothgrad}, and LIME.

Attribution methods have also been criticized for confirmation bias \cite{ancona2017towards}. An appealing but incorrect explanation might be judged more favorably than a more realistic one. A better understanding of the goals of an idealized attribution method is needed to develop improved quantitative tools for XAI evaluation \cite{ancona2017towards}. In \cite{adebayo2021post}, the limitations of post-hoc explanations are investigated. The authors question their effectiveness in detecting unknown (to the user at test time) spurious correlations. These inefficiencies are detected in three types of post-hoc explanations: feature attribution, concept activation, and training point ranking. However, the authors acknowledge that these three classes do not fully cover all post-hoc explanation methods. Other methods have been criticized for their weak or untrustworthy causal relationships. In \cite{atrey2019exploratory}, saliency maps are criticized for their frequent unfalsifiability and high subjectivity. The study also highlights their causal unreliability in reflecting semantic concepts and agent behavior in reinforcement learning environments. In \cite{nguyen2021effectiveness}, it is argued that feature attribution techniques are not more effective than showing the nearest training-set data point when tested on humans. The limitations of attribution methods in cases of non-visible artifacts \cite{zhou2022feature} have also been investigated.

Despite the critical studies on explainable classification and their potential extensions to segmentation, the widespread prevalence of image segmentation requires investigating different explainability tools and their working mechanisms. Although some studies point out the limitations of these techniques, better alternatives have yet to be developed. As observed in \cite{geirhos2023don}, the development of interpretability methods is dialectical: a new method is introduced, its failure modes are identified, and as a result, a new method is proposed,  with the ongoing aim of making them more reliable. Current methods have much room for improvement, especially considering that the entire field is in the early stages of development. The above criticisms can serve as sanity checks for XAI methods. Despite the limitations, some techniques, such as gradients and Grad-CAM in the case of \cite{adebayo2018sanity}, do pass certain sanity checks. Even some critical literature \cite{geirhos2023don} agrees that certain explainability techniques can be useful for exploratory use cases. To our knowledge, the specifics of XAI limitations in image segmentation have not yet been explored.

\section{Taxonomy}

Different XAI taxonomies have been introduced in classification, both with respect to specific subgroups of interpretability methods \cite{arrieta2020explainable}, \cite{fan2021interpretability}, and with respect to more abstract conceptual terms \cite{graziani2023global}. Even meta-reviews of various existing taxonomies have been proposed \cite{speith2022review}, \cite{schwalbe2023comprehensive}. Since image segmentation can be seen as an extension of classification, many taxonomy-related aspects can be validly transferred from research in explainable classification. In most taxonomies, a particularly important role is played by three dichotomies: post-hoc vs ad-hoc (sometimes also referred to as inherent interpretability), model-specific vs model-agnostic, and local vs global explanations.

\subsection{Scope: Local vs Global}
The first prevalent dichotomy distinguishes between local and global explanations. Here, locality refers to the use of a single input image with respect to which the explanation is given. A global explanation, on the other hand, would aim to explain the model's behavior across a range of different images, not limiting itself to just one. According to meta-surveys \cite{speith2022review}, \cite{schwalbe2023comprehensive}, the local-global dichotomy is prevalent in numerous XAI taxonomies. This distinction is essential in explainable segmentation as well, with most methods falling under local explanations.

\subsection{Method and its timing: Post-hoc vs Ad-hoc}

The distinction between post-hoc and ad-hoc explanations highlights that one can either apply XAI techniques to an already-trained model without any interference or apply them during and as part of the training process. Sometimes, these explanations are also described as passive and active approaches \cite{zhang2021survey}. Under this definition, active approaches require modifications to the network or the training process. Such changes influence both the model's performance in terms of evaluative metrics and its interpretability. Therefore, an accuracy-interpretability trade-off cannot be avoided in ad-hoc XAI methods, but it is avoided in the case of post-hoc applications.

This widely accepted dichotomy can nonetheless be slightly misleading, as both terms can be meant to emphasize different distinctive criteria. Post-hoc can be understood as referring to the fact that XAI techniques are applied after the training, hence ``post''. Naturally, it would seem that ad-hoc should be understood as referring to XAI techniques that are applied during training. However, sometimes, as a direct opposition to ``post-hoc'', terms like 'inherent interpretability' \cite{rudin2019stop} or ``self-explainability'' \cite{sacha2023protoseg} are used, pointing to an entirely different aspect: the architecture or the type of XAI method. In some cases, such interpretation could allow for XAI methods that are both inherently interpretable and post-hoc \cite{molnar2020interpretable}, which might cause confusion.

\subsection{Range: Model-specific vs Model-agnostic}
The third distinction evaluates the flexibility of a given XAI technique in its application to different model architectures. Model-specific XAI methods heavily depend on the underlying model architecture, whereas model-agnostic methods are more universal in their compatibility with various models, and can be applied to different architectures without further modifications. The interpretation of inherently interpretable models is always model-specific \cite{molnar2020interpretable}.

\subsection{XAI taxonomy for image segmentation}
\begin{figure}
        \centering
        \includegraphics[width=0.9\linewidth]{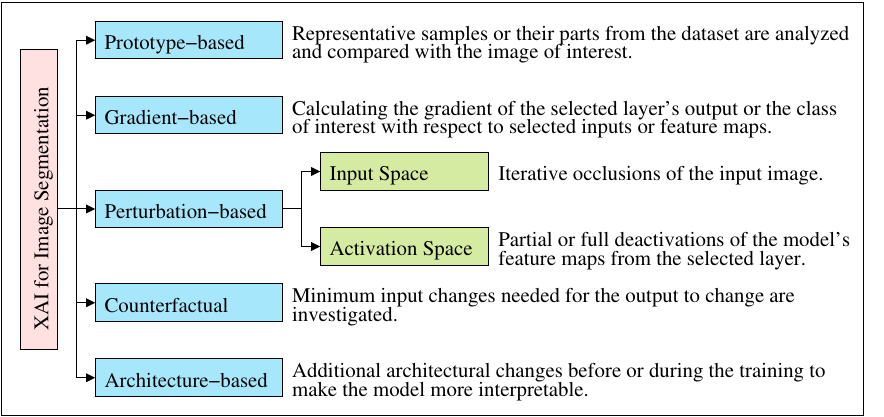}
        \caption{Method-centered taxonomy}
        \label{fig:enter-label2}
\end{figure}

Multiple compatible taxonomies are possible depending on the level of abstraction in which we are interested. In \cite{arrieta2020explainable}, XAI methods in ML are divided into transparent models and post-hoc explainability, which is then further divided into model-specific and model-agnostic categories. In \cite{fan2021interpretability}, interpretation methods are divided into post-hoc interpretability analysis and ad-hoc interpretable modeling. In \cite{shahroudnejad2021survey}, a higher-level taxonomy distinguishes between structural analysis, behavioral analysis, and explainability by design. In \cite{chromik2020taxonomy}, a preliminary taxonomy of human subject evaluation in XAI is introduced, which might be particularly useful when using qualitative evaluations of XAI. Based on the analysis of XAI taxonomies in classification, we observe that they could also be applied to image segmentation. However, no specific framework has been introduced to address the ever-growing field of interpretable segmentation. We hope that a more detailed demarcation will be useful in navigating across different types of techniques.  In our survey, we propose a taxonomy (Fig. \ref{fig:enter-label2}) that is based on the reviewed literature in explainable image segmentation.

The proposed method-centered taxonomy includes five method families: prototype-based, gradient-based, perturbation-based, counterfactual methods, and architecture-based techniques. Prototype-based methods employ representative samples or their parts from the dataset to analyze and compare with the input image. Gradient-based methods involve calculating the gradient of the output of a selected layer or the class of interest with respect to selected inputs or feature maps. Perturbation-based methods can be divided into two groups based on the perturbed space. Input space perturbations are iterative occlusions of the input image. Typically, they are based on a sliding filter, but different types of noise can also be introduced. Explanations are based on their effect on the model's outcome. Activation space perturbations involve partial or full deactivations of the model's feature maps from the selected layer. Once again, explanations are based on their effect on the model's outcome. Counterfactual methods employ the minimum input changes needed for the output to change. Finally, architecture-based techniques involve making additional architectural changes either before or during training to enhance interpretability. Section 4 presents a more detailed analysis of each method group.

\section{XAI for Image Segmentation}

In this section, we review the main methods representative of each subgroup in the taxonomy, as well as the metrics for explainable image segmentation.

\subsection{Methods}

\subsubsection{Prototype-based methods}

Prototype-based models \cite{biehl2016prototype} utilize typical representatives from the dataset, usually selected from the training set. These methods emphasize the intuitiveness of the provided explanations, presenting them in an easily understandable form of naturally occurring objects. Such features can be easily distinguished and discriminated by end users. Meanwhile, prototypical parts refer to specific regions within representative prototypes, also known as exemplars. In contrast to a prototype, a criticism is a data instance that is not well represented by the prototypes \cite{molnar2020interpretable}. In terms of architecture, typical prototype-based methods require the insertion of a prototype layer into the segmentation model. Therefore, depending on the taxonomy, prototype-based methods could also be viewed as self-explainable and part of the architecture-based methods. However, due to their frequent mentions in the related classification literature under the same subgroup label, we opt to treat them as a separate group. 

\begin{figure*}[tbh]
 \centering
 \resizebox{0.99\textwidth}{!}{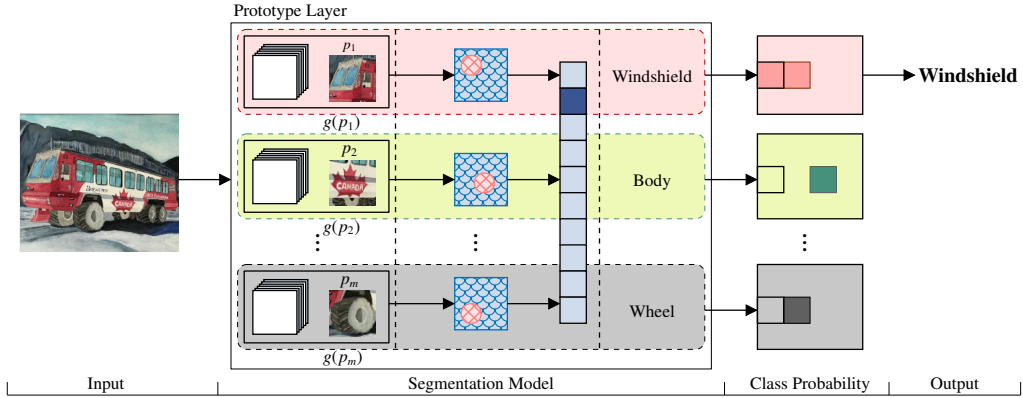}
 \caption{A framework for prototype-based methods.}
 \label{fig:prototype}
\end{figure*}

Although prototypical methods are prevalent in classification \cite{chen2019looks}, \cite{donnelly2022deformable}, \cite{rymarczyk2022interpretable}, their extensions for segmentation are few. Typically, prototype layer (Fig. \ref{fig:prototype}) is a key component in prototype-based methods for both classification \cite{chen2019looks}, \cite{donnelly2022deformable}, \cite{rymarczyk2022interpretable} and segmentation \cite{zhang2022interpretable}, \cite{sacha2023protoseg}. Within a prototypical layer, different classes are represented by predefined or learned prototypes. In \cite{sacha2023protoseg}, a ProtoSeg model is proposed. The authors introduce a diversity loss function based on Jeffrey's divergence \cite{jeffreys1998theory} to increase the prototype variability for each class. Better results are observed when the diversity loss component is introduced. The authors attribute this to the higher informativeness of a more diverse set of prototypes that leads to a better generalization. We think that this could be related to the diversity hypothesis \cite{hilton2020understanding}, first introduced in the context of reinforcement learning, and could be explored further. The experiments are performed using Pascal VOC 2012 \cite{everingham2010pascal}, Cityscapes \cite{cordts2016cityscapes}, and EM Segmentation Challenge \cite{imagejSegmentationNeuronal} datasets. DeepLab \cite{chen2017deeplab} model is used as the backbone. In \cite{zhang2022interpretable}, a prototype-based method is used in combination with if-then rules for the interpretable segmentation of Earth observation data. The proposed approach is the extension of xDNN \cite{angelov2020towards}, and uses mini-batch K-mean \cite{sculley2010web} clustering. For the feature extraction part, the U-Net architecture is used. The experiments are performed using the Worldfloods \cite{mateo2021towards} dataset.

\subsubsection{Counterfactual explanations}

\begin{figure*}[tbh]
 \centering
 \resizebox{0.99\textwidth}{!}{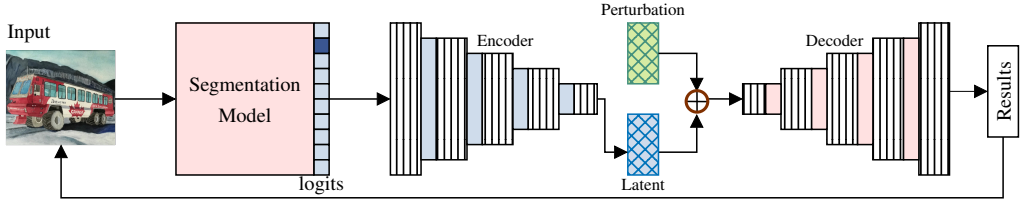}
 \caption{A framework for counterfactual methods.}
 \label{fig:Counterfactual}
\end{figure*}

Counterfactual or contrastive explanations investigate the minimum input changes needed for the output to change. Unconditional counterfactual explanations were first introduced in \cite{wachter2017counterfactual}. This explainability subfield is related to adversarial attacks. Counterfactual images are similar to the original, yet are able to change the model's output. Counterfactual explanations can also be viewed as closely linked to perturbation-based explanations, which will be discussed in the next subsection. Counterfactual XAI techniques frequently fall into local post-hoc category \cite{guidotti2022counterfactual}. After the initial segmentation model, counterfactual-based interpretability method typically employs additional networks for counterfactual generation. In our pipeline (Fig.~\ref{fig:Counterfactual}), this is depicted by additional encoder and decoder networks. 

Generator-based counterfactual explanations are investigated in \cite{zemni2023octet}. OCTET, a generative approach, produces object-aware counterfactual explanations for complex scenes.  Counterfactual changes to the image focus on road markings, such as changing the solid line into a dashed one, or the positions of cars by cropping and extending the relevant regions of the input image. The models are trained on the BDD100k \cite{yu2020bdd100k} and BDD-OIA \cite{xu2020explainable} datasets.
Additional information can be found in the supplementary material \cite{zemnioctet}. In \cite{singh2022counterfactual}, segmentation results are qualitatively compared using counterfactual images. The experiments are performed on Kvasir-seg \cite{jha2020kvasir} and Kvasir-instrument \cite{jha2021kvasir} datasets. Counterfactual explanations are generated using the segmented area of interest, which is then replaced with the average pixel value of the rest of the image. In \cite{jacob2022steex}, counterfactual explanations are generated for complex scenes while preserving the semantic structure. The proposed method uses semantic-to-real image synthesis. Here, a noticeable contrast can be drawn between this approach and perturbation-based methods. In the latter, perturbations applied to the input space fail to produce semantically meaningful image regions.

\subsubsection{Perturbation-based methods}

\begin{figure}[tbh]
 \centering
 \resizebox{0.7\textwidth}{!}{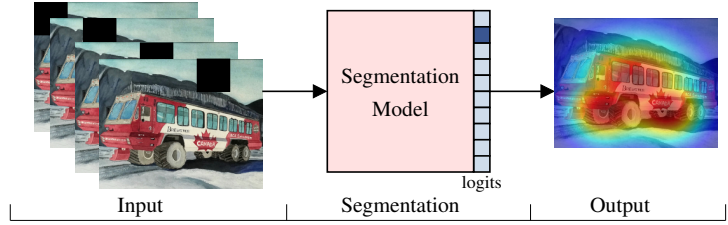}
 \caption{A framework for perturbation-based methods for the input space.}
 \label{fig:inputspace}
\end{figure}

\begin{figure}[tbh]
 \centering
 \resizebox{0.7\textwidth}{!}{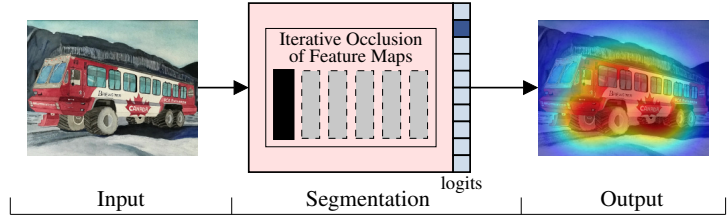}
 \caption{A framework for perturbation-based methods for the activation space.}
 \label{fig:activation}
\end{figure}

Perturbation-based methods typically employ occlusions in the input (Fig.~\ref{fig:inputspace}) or activation (Fig.~\ref{fig:activation}) space, and then measure their influence on the model's output for a selected class. Here occlusions or perturbations could be understood as uninformative regions, transforming the input or its internal feature maps. Pixels in the occlusion filter can be set to 0, as seen in Fig. \ref{fig:inputspace}, or any other arbitrary value. Such sliding filter would occlude different regions of the input space. Its size and stride parameters are specified beforehand. Gaussian or any other random noise can also be used for these purposes. Multiple perturbative iterations are required for the generation of an explanation map. During each inference, the score is calculated, measuring the perturbation's effect on model's performance. This can be done by taking the difference between the score for the original image and that of its perturbed version. Such a score can be based on an evaluative metric or pre-Softmax prediction values. Since the same input image has to undergo multiple transformations, each requiring a separate forward pass through the model, perturbation-based XAI methods are considered computationally expensive. 

Following \cite{gipivskis2023ablation}, given an RGB image $x$ of dimensions \textit{$N \times M \times 3$}, the pre-Softmax prediction score for pixel $x_{ij}$ for class $c$ is $ l^{c}(x_{ij})$. When $x_{ij}$ is classified as $c$, the sum of these scores is:

\begin{equation}
L^{c}(x) = \sum_{i, j} [\hat{c}_{ij} = c]l^{c}(x_{ij}),
\end{equation}

\noindent
where $\hat{c}_{ij}$ is the predicted class for ${x_{ij}}$.

Following \cite{ramaswamy2020ablation}, $L^{c}(x)$ is used to compute the importance weight $w_k^c$ of each activation map $k$: 

\begin{equation}
w_k^c = \frac{L^{c}(x) - L_k^{c}(x)}{L^{c}(x)},
\end{equation}

\noindent
where $L_k^{c}(x)$ is the sum of pre-Softmax scores for $c$ after deactivating $k$. These calculated importance scores are then used in a linear combination of feature maps to generate a perturbation-based explanation.

The authors of \cite{hoyer2019grid} propose the first XAI solution for extending saliency techniques beyond classification. Their perturbation-based method is introduced for the detection of contextual biases. The experiments are performed using a synthetic toy dataset based on MNIST \cite{lecun1998mnist} as well as the Cityscapes \cite{cordts2016cityscapes} dataset. In \cite{wan2020segnbdt}, a hybrid SegNBDT approach is introduced, combining both decision trees and neural networks. This method falls under both the perturbation-based and self-explainable model categories. For the experimental part, Pascal Context \cite{mottaghi2014role}, Cityscapes \cite{cordts2016cityscapes}, and Look Into Person \cite{gong2017look} datasets are used. In \cite{dardouillet2022explainability}, SHAP and RISE techniques are applied to image segmentation. SHAP is a popular post-hoc interpretability method, and the proposed approach is based on Kernel SHAP \cite{lundberg2017unified}. The experiments are performed on Synthetic Aperture Radar images from the unspecified dataset for oil slick detection at the sea surface and the Cityscapes \cite{cordts2016cityscapes} dataset. 
In \cite{koker2021u}, a perturbation-based occlusion sensitivity approach is used to measure the performance of the proposed interpretable semantic segmentation approach. Compared to occlusion sensitivity and Grad-CAM, their method achieves orders of magnitude lower inference time. However, it requires training an additional interpretability model. In \cite{gipivskis2023occlusion}, following \cite{zeiler2014visualizing}, different types of input occlusions are investigated for applications in semantic segmentation. The paper discusses how occlusion filter sizes and colors can affect the generated explanations. It is observed that, compared to image classification, input occlusions in segmentation models do not generate as much variance in the evaluation metric scores. For the experimental investigation, COCO \cite{lin2014microsoft} dataset is used. The proposed method is evaluated qualitatively, with select images also compared quantitatively using deletion curves.

Perturbations are not limited to the input space. For instance, Ablation-CAM \cite{ramaswamy2020ablation} is a gradient-free method that systematically deactivates feature maps in a selected layer. In \cite{gipivskis2023ablation}, Ablation-CAM  is extended to semantic segmentation. It is a gradient-free interpretability technique based on ablating or perturbing activation maps. The experiments are performed on a private industrial dataset for fruit-cutting machines as well as on COCO \cite{lin2014microsoft} dataset.

\subsubsection{Gradient-based methods}

\begin{figure*}[tbh]
 \centering
 \resizebox{0.8\textwidth}{!}{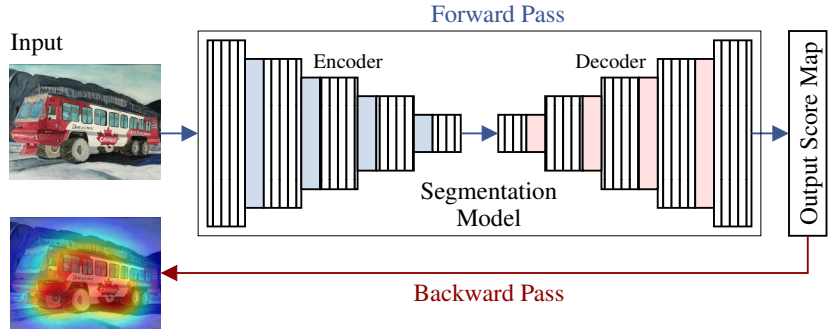}
 \caption{A framework for gradient-based methods.}
 \label{fig:gradient}
\end{figure*}

Gradient-based methods (Fig. \ref{fig:gradient}) typically use gradients of the outputs from later layers with respect to the input features. These techniques are less computationally expensive compared to perturbation-based techniques because only a single backward pass is required. Perturbation techniques, on the other hand, require a separate forward pass for each perturbed image, increasing computational costs with each inference.

Following \cite{gipivskis2023impact}, given an RGB image $x$ of dimensions \textit{$N \times M \times 3$}, and a set of class labels $\{1, 2, ..., C\}$, where C denotes the total number of classes, we define:

\begin{equation}
g(x) = (g_{1}(x_{ij}), ..., g_{C}(x_{ij})) \in \mathbb{R}^{N \times M \times C},   
\end{equation}
where $g_{c}(x_{ij})$ denotes the pre-Softmax prediction score for class $c$ at pixel $x_{ij}$.

The sum of these pre-Softmax scores for class $c$ is then defined as: 

\begin{equation}    
g_{c, A}(x) = \sum_{i, j \in A} g_{c}(x_{ij}),
\end{equation}
where $A$ is a set of pixel indices of interest.

From this, the gradient-based explanation with respect to $c$ is derived as: 
\begin{equation}
    G_{A}(x, c) = \frac{\partial  g_{c, A}(x)}{\partial x}.
\end{equation}

In \cite{vinogradova2020towards}, 
Seg-Grad-CAM is proposed as the extension of Grad-CAM \cite{selvaraju2017grad}. It is one of the best known explainability techniques in image segmentation. Just like in the case of regular Grad-CAM, the generated saliency is based on the weighted sum of the selected feature maps. Its application is demonstrated on a U-Net model, trained on the Cityscapes \cite{cordts2016cityscapes} dataset. In \cite{chiu2023potential}, the same method is applied for automatic rock joint trace mapping. The original Grad-CAM technique for classification, together with simple gradients, passes the previously discussed sanity checks, evaluating the reliability of XAI technique. In \cite{hasany2023seg}, Seg-XRes-CAM is introduced. The authors criticize Seg-Grad-CAM \cite{vinogradova2020towards} for not utilizing spatial information when generating saliency maps for a region of the segmentation map. The proposed approach draws inspiration from HiResCAM \cite{draelos2020use}, a modification of the original Grad-CAM \cite{selvaraju2017grad}. Subsequently, \cite{gizzini2023extending} adapts five CAM-based XAI methods from classification to the segmentation of high-resolution satellite images. Among the proposed extensions are Seg-Grad-CAM++, Seg-XGrad-CAM, Seg-Score-CAM, and Seg-Eigen-CAM. Just like in \cite{gipivskis2023ablation}, Ablation-CAM, a gradient-free method, is also extended for segmentation. Besides using the drop in the segmentation score to measure their methods' performance, the authors also propose entropy-based XAI evaluation metric. The implemented methods are tested on a WHU \cite{ji2018fully} building dataset. In \cite{schorr2021neuroscope}, an interpretability and visualization toolbox is proposed for classification and segmentation networks. It includes several XAI extensions specifically for image segmentation. Among them are Guided Grad-CAM and segmented score mapping, extended from \cite{kapishnikov2019xrai}.

\subsubsection{Architecture-based methods}

\begin{figure}[tbh]
 \centering
 \resizebox{0.7\textwidth}{!}{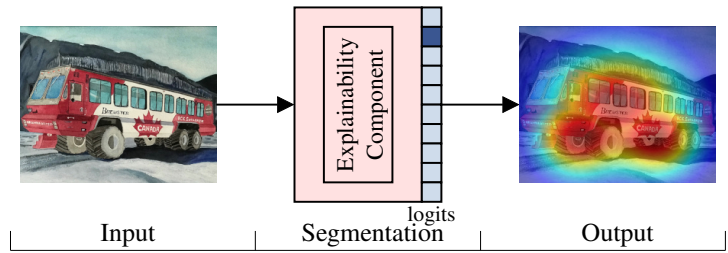}
 \caption{A framework for architecture-based methods.}
 \label{fig:architecture}
\end{figure}

This subgroup of methods introduces additional architectural changes (Fig.~\ref{fig:architecture}) that aim to make the models more interpretable. Instead of relying on post-hoc techniques that are added on top of the already trained models, these methods are typically employed as part of the training process. This class of XAI methods is sometimes described as interpretable by design, inherently interpretable, or interpretability as part of the architecture.

One such example is chimeric U-Net with an invertible decoder \cite{schulze2022chimeric}. This approach introduces architectural constraints for the sake of explainability. The authors claim that it can achieve both local and global explainability. In \cite{losch2021semantic},  both supervised and unsupervised techniques of Semantic Bottlenecks (SB) are introduced for better inspectability of intermediate layers. This approach is proposed as an addition to the pre-trained networks. Unsupervised SBs are identified as offering greater inspectability compared to their supervised counterparts. The experiments are primarily performed on street scene segmentation images from the Cityscapes dataset. The results are also compared using two other datasets: Broden \cite{bau2017network} and Cityscapes-Parts, a derivative of Cityscapes. In \cite{santamaria2020towards}, a framework for symbolic semantic segmentation is proposed. This work is at the intersection of image segmentation and emergent language models. The authors apply their research to medical images, specifically brain tumor scans. Emergent Language model with a Sender and a Receiver is utilized for interpretable segmentation. The Sender is an agent responsible for generating a symbolic sentence based on the information from the high model layer, while the Receiver cogenerates the segmentation mask after receiving symbolic sentences. Symbolic U-Net is trained on the Cancer Imaging Archive (TCGA) dataset\footnote{https://wiki.cancerimagingarchive.net/pages/viewpage.action?pageId=5309188} and used for providing inputs to the Sender network. 

\subsection{Metrics}

XAI techniques are used in addition to standard evaluation metrics due to their limitations. However, to evaluate the performance of these techniques, they also need to be measured. We can distinguish between qualitative and quantitative assessment methods. Qualitative evaluation commonly refers to user-based evaluation and, based on the surveyed papers (Table \ref{tab:medical} and Table \ref{tab:industry}), is the more prevalent of the two. To quantify subjective user results, various questionnaires have been proposed \cite{hoffman2018metrics}, such as the explanation goodness checklist, explanation satisfaction scale, trust scales, and the ease of understanding when comparing different explainability techniques \cite{graziani2021sharpening}. These methods still require polling multiple subjects, although, when surveying experts, in practice their number is limited to 2-5 \cite{graziani2021sharpening}. This way, quantification still takes place, but it is based on subject-dependent evaluation. Since questionnaire studies require additional resources, most of the papers using qualitative evaluation only provide visual comparisons between different XAI techniques, leaving qualitative evaluation to the reader's eye. 

Quantitative evaluation does not involve human subjects and can be more easily applied when comparing different interpretability methods. Infidelity and sensitivity \cite{yeh2019fidelity} are the only two metrics that, as of 2024, are implemented in the Captum \cite{kokhlikyan2020captum} interpretability library for PyTorch. Deletion and insertion metrics \cite{petsiuk2018rise} are another type of quantitative evaluation, based on measuring the Area under the Curve (AUC), generated after gradually deleting or inserting the most important pixels in the input space. However, for some XAI methods, such as counterfactual explanations, it might be difficult to evaluate the usefulness of the explanation quantitatively. In the case of counterfactual explanations, we can measure whether the generated images are realistic and how closely they resemble the query images, but for a more thorough evaluation of the explanation itself, user studies  \cite{zemni2023octet} might be required.

In \cite{colin2022cannot}, a psychophysics study (\textit{N} = 1,150) is conducted to evaluate the performance of six explainable attribution methods on different neural network architectures. Shortcomings in the methods are detected when using them to explain failure cases. Comparative quantitative rankings of different saliency techniques can also be inaccurate. In \cite{tomsett2020sanity}, inspired by \cite{adebayo2018sanity}, sanity checks for saliency maps are investigated. The authors perform checks for inter-rater reliability, inter-method reliability, and internal consistency, and determine that the current saliency metrics are unreliable. It is observed that these metrics exhibit high variance and are sensitive to implementation details.

\section{Applications}

This section presents concrete XAI applications in medical and industrial domains. We also discuss other use cases, with a primary focus on industry-related monitoring domains, such as remote sensing, environmental observation, and biometrics. Additionally, the potential uses of XAI for self-supervised image segmentation are reviewed.

\subsection{Medical Applications}

\begin{table*}[htb]
  \newcommand{\z}{\phantom{0}}
  \newcommand{\za}{\phantom{,}}
  \newcolumntype{m}{>{\hsize=.7\hsize}X}
  \newcolumntype{b}{>{\hsize=.3\hsize}X}
  \newcolumntype{s}{>{\hsize=.2\hsize}X}
  \newcolumntype{q}{>{\hsize=1.6\hsize}X}
  \newcolumntype{L}{>{\raggedright}X}
  \centering
  \scriptsize
  \caption{Explainable Image Segmentation in Medicine}
\begin{tabularx}{\textwidth}{b|mqqssss} \hline 
Field	&	Imaging modality	&	Objects of interest	&	Datasets	& Metric & Year &	Ref.	\\\hline
G	&	IMG*	&	Colorectal polyps	&	EndoScene \cite{vazquez2017benchmark}	& $\vartriangleright$ &	2018	&	\cite{wickstrom2020uncertainty}	\\\hline
O	&	CT	&	Liver tumors	&	LiTS \cite{bilic2023liver}	& $\blacktriangleright$ &	2019	&	\cite{couteaux2019towards}	\\\hline
C	&	CMRI	&	Ventricular volumes	&	SUN09 \cite{radau2009evaluation}, AC17 \cite{bernard2018deep}	& $\vartriangleright$ &	2020	&	\cite{sun2020saunet}	\\\hline
O	&	MRI	&	Brain tumors	&	TCGA	& $\vartriangleright$ &	2020	&	\cite{santamaria2020towards}	\\\hline
D	&	MRI/IMG	&	Skin lesions, multi-organ (incl. the fetal brain and the placenta)	&	ISIC2018 \cite{codella2019skin} and a private fetal MRI dataset	& $\vartriangleright$ &	2020	&	\cite{gu2020net}	\\\hline
P	&	CT	&	Pancreatic region	&	Medical segmentation decathlon	& $\blacktriangleright$ &	2021	&	\cite{koker2021u}	\\\hline
C	&	CMRI	&	Ventricles, myocardium	&	Cardiac MRI dataset \cite{bernard2018deep}	& $\vartriangleright$ &	2021	&	\cite{janik2021interpretability}	\\\hline
G	&	IMG	&	Polyps, med. instruments	&	Kvasir-SEG \cite{jha2020kvasir}, Kvasir-Instrument \cite{jha2021kvasir}	& $\vartriangleright$ &	2021	&	\cite{ahmed2021explainable}	\\\hline
O	&	MRI	&	Brain tumors	&	BraTS2018 \cite{menze2014multimodal}	& $\blacktriangleright$ &	2021	&	\cite{saleem2021visual}	\\\hline
FM	&	NIR	&	Iris	&	Private test dataset, post-mortem iris datasets, collected by \cite{trokielewicz2020post}	& $\vartriangleright$ &	2022	&	\cite{kuehlkamp2022interpretable}		\\\hline
V	&	CT/MRI/IMG	&	Skin lesions, abdomen multi-organ, brain tumors	&	HAM10000 \cite{tschandl2018ham10000}, CHAOS 2019 \cite{kavur2021chaos}, BraTS 2020 \cite{menze2014multimodal}	& $\vartriangleright$ &	2022	&	\cite{karri2022explainable}	\\\hline
V	&	MRI	&	Brain tumors, human knees	&	BraTS 2017 \cite{menze2014multimodal}, OAI ZIB \cite{ambellan2019automated}	& $\vartriangleright$&	2022	&	\cite{schulze2022chimeric}	\\\hline
O	&	MRI	&	Brain tumors&	BraTS 2019 \cite{menze2014multimodal}, BraTS 2021 \cite{menze2014multimodal}	& $\vartriangleright$&	2022	&	\cite{zeineldin2022explainability}	\\\hline
N &	MRA	&	Brain vessels &	Private	& $\blacktriangleright$&	2022	&	\cite{chatterjee2022torchesegeta}	\\\hline
H	&	IMG	&	Liver	&	 Simulated dataset (Test Set 4) \cite{bardozzo2022stasis}	& $\vartriangleright$&	2022	&	\cite{bardozzo2022cross}	\\\hline
O	&	US/MG &	Breast tumors	&	Private LE/DES datasets, and BUSI \cite{al2020dataset}	& $\vartriangleright$ &	2023	&	\cite{wang2023information}	\\\hline
G	&	CT/IMG	&	Colorectal polyps, lung cancer	&	EndoScene \cite{vazquez2017benchmark}, LIDC-IDRI \cite{armato2011lung}	& $\blacktriangleright$ &	2023	&	\cite{cheng2023tax}	\\\hline
O	&	CT/MRI	&	Prostate cancer	&	3D pelvis dataset \cite{dowling2015automatic}	& $\vartriangleright$ &	2023	&	\cite{dai2023explainable}	\\\hline
G	&	CT	&	Abdominal organs	&	Synapse multi-organ CT dataset \cite{landman2015miccai}	& $\vartriangleright$ &	2023	&	\cite{hasany2023seg}	\\\hline
O	&	BUS	&	Breast tumors	&	BUSI \cite{al2020dataset}, BUSIS \cite{xian2018benchmark}, HMSS \cite{geertsma2014ultrasoundcases}	& $\vartriangleright$ &	2023	&	\cite{karimzadeh2023post}	\\\hline
O	&	CT/PET	&	Non-small cell lung cancer, whole-body	&	NSCLC, AutoPET \cite{gatidis2022whole}	& $\blacktriangleright$ &	2023	&	\cite{kang2023learning}	\\\hline
D	&	IMG	&	Skin lesions	&	ISIC2018 \cite{codella2019skin}	& $\vartriangleright$ &	2023	&	\cite{wang2023explainable}
	\\\hline
D	&	IMG	&	Melanoma	&	ISIC 2018 \cite{codella2019skin}, ISIC 2019 \cite{tschandl2018ham10000}	& $\vartriangleright$ &	2023	&	\cite{sun2023explainable}
\\\hline
O	&	MRI/IMG	&	Prostate tumors, optic disc and cup	&	Prostate** and fundus*** datasets	& $\vartriangleright$ &	2023	&	\cite{zhang2023s}	\\\hline
O	&	X-ray	&	Breast tumors	&	INbreast \cite{moreira2012inbreast}	& $\blacktriangleright$ &	2023	&	\cite{farrag2023explainable}	\\\hline
O	&	WSI	&	Head and neck tumors	&	Private	& $\vartriangleright$ &	2023	&	\cite{dorrich2023explainable}	\\\hline
A	&	IR	&	Feet	&	ThermalFeet	& $\vartriangleright$ &	2023	&	\cite{aguirre2023feet}	\\\hline
V	&	CT/MRI/IMG	&	Brain tumors	&	BraTS 2018 \cite{menze2014multimodal}, BraTS 2019 \cite{menze2014multimodal}, BraTS 2020 \cite{menze2014multimodal}, ISIC 2017	& $\vartriangleright$ &	2023	&	\cite{he2023segmentation}	\\\hline
P	&	CT	&	Pancreas	&	Pancreas segmentation dataset \cite{antonelli2022medical}	& $\vartriangleright$ &	2023	&	\cite{okamoto2023generating}	\\\hline
Op	&	OCT	&	Retinal layers, glaucoma, diabetic macular edema	&	NR206, glaucoma dataset \cite{li2021multi}, DME dataset \cite{chiu2015kernel}	&  $\vartriangleright$ &	2023	&	\cite{he2023exploiting}	\\\hline
V	&	CT/MRI	&	Prostate, left ventricle, right ventricle, myocardium	&	NCI-ISBI 2013 \cite{bloch2015nci}, I2CVB \cite{lemaitre2015computer}, PROMISE12 \cite{litjens2014evaluation}; MSCMR \cite{zhuang2018multivariate}, EMIDEC \cite{lalande2020emidec}, ACDC \cite{bernard2018deep}, MMWHS \cite{zhuang2016multi}, CASDC 2013 \cite{kiricsli2013standardized}	& $\vartriangleright$ &	2023	&	\cite{gao2023bayeseg}	\\\hline
Op	&	OCT	&	Retinal layers, glaucoma, diabetic macular edema	&	Vis-105H, glaucoma dataset \cite{li2021multi}, DME dataset \cite{chiu2015kernel}	&  $\vartriangleright$ &	2024	&	\cite{he2024light}	\\\hline
V	&	CMRI/CT	&	Left atrium, thoracic organs	&	Atrium dataset \cite{antonelli2022medical}, SegTHOR \cite{lambert2020segthor}	&  $\vartriangleright$ &	2024	&	\cite{lambert2024incorporation}	\\\hline

\multicolumn{6}{l}{A: Anesthesiology, C: Cardiology, D: Dermatology, FM: Forensic Medicine, G: Gastroenterology, N: Neurology,}\\
\multicolumn{6}{l}{ O: Oncology, Op: Ophthalmology, P: Pancreatology, and V: Various}\\
\multicolumn{6}{l}{*  IMG: general-purpose digital image formats, such as JPEG}\\ 
\multicolumn{6}{l}{**  Prostate datasets: RUNMC \cite{bloch2015nci}, BMC \cite{bloch2015nci}, HCRUDB \cite{lemaitre2015computer}, UCL \cite{litjens2014evaluation}, BIDMC \cite{litjens2014evaluation}, and HK \cite{litjens2014evaluation} }\\ 
\multicolumn{6}{l}{*** Fundus datasets: DRISHTI-GS \cite{sivaswamy2015comprehensive}, RIM-ONE-r3 \cite{fumero2011rim}, and REFUGE \cite{orlando2020refuge}}\\
\multicolumn{6}{l}{$\vartriangleright$: Qualitative XAI evaluation}\\
\multicolumn{6}{l}{$\blacktriangleright$:  Quantitative XAI evaluation}\\

    \end{tabularx}
   
\label{tab:medical}
\end{table*}

Most applications in explainable image segmentation have been investigated in the medical domain, using datasets from various medical fields (Table \ref{tab:medical}), ranging from cardiology to oncology. Proposed XAI solutions and applications are employed for diagnosing, monitoring, and other clinical tasks. In some cases, there might be unavoidable overlaps between medical fields. For instance, overlaps occur at the intersection of oncology and histopathology when discussing microscopic tumor images, or between oncology and dermatology when considering melanoma \cite{sun2023explainable}. Such overlaps can also arise from using multiple datasets, each associated with a different medical field. In these instances, we specify the relevant details in the method description.

\subsubsection{Dermatology}

Dermatology-centered XAI applications \cite{sun2023explainable}, \cite{wang2023explainable} focus on skin lesions. Specifically, \cite{sun2023explainable} discusses applications for interpreting melanoma diagnosis results. The proposed pipeline utilizes both classification and segmentation networks. Grad-CAM is employed to generate explainable heatmaps for the classifier, which are then used as inputs in the U-Net network. These heatmaps assist in generating indicator biomarker localization maps. The proposed approach can be used in self-supervised learning. Experiments are performed on the ISIC 2018 \cite{codella2019skin} and ISIC 2019 \cite{tschandl2018ham10000}, \cite{codella2018skin}, \cite{combalia2019bcn20000} datasets. In \cite{wang2023explainable}, a CAM-based explainability metric is proposed and incorporated into the loss function. This metric quantifies the difference between the CAM output and the segmentation ground truth for the targeted class. Both segmentation and explanation losses are considered during the model's training phase. The use of CAM with learnable weights enables a balance between segmentation performance and explainability. The proposed method belongs to the self-explainable XAI category. Similar to \cite{sun2023explainable}, the U-Net network is used. The experiments are conducted on the ISIC2018 \cite{codella2019skin} dataset. In \cite{gu2020net}, a comprehensive attention-based
convolutional neural network is proposed for better interpretability in dermoscopic and fetal MRI images. This approach uses multiple attentions, combining the information about spatial regions, feature channels, and scales. The experiments are performed on ISIC 2018 \cite{codella2019skin} and a private fetal MRI dataset.

\subsubsection{Forensic Medicine}

The applications of explainable segmentation in forensic medicine are limited to iris segmentation. This can be more narrowly referred to as forensic ophthalmology. In \cite{kuehlkamp2022interpretable}, the investigation focuses on forensic postmortem iris segmentation. The authors apply a classical technique of Class Activation Mapping (CAM) \cite{zhou2016learning}. The experiments are performed on a private test dataset, and publicly available post-mortem iris datasets, collected by \cite{trokielewicz2020post}.

\subsubsection{Gastroenterology}

XAI applications for endoscopic image segmentation primarily focus on polyps. In \cite{wickstrom2020uncertainty}, the Guided Backpropagation \cite{springenberg2014striving} technique is extended to the semantic segmentation of colorectal polyps. Uncertainty in input feature importance is estimated, with higher uncertainty observed in inaccurate predictions. Uncertainty maps are generated using the Monte Carlo Dropout method. The proposed solution is evaluated on the EndoScene \cite{vazquez2017benchmark} dataset. In \cite{ahmed2021explainable}, Layer-wise Relevance Propagation (LRP), a propagation-based explainability method, is applied to the endoscopic image segmentation of gastrointestinal polyps and medical instruments. LRP is specifically applied to the generator component within a Generative Adversarial Network. The generated relevance maps are then qualitatively evaluated. The segmentation models are trained on the Kvasir-SEG \cite{jha2020kvasir} and Kvasir-Instrument \cite{jha2021kvasir} datasets. 

\subsubsection{Hepatology}

In \cite{bardozzo2022cross}, two gradient-based post-hoc explanations, Grad-CAM and Grad-CAM++, are investigated for cross explanation of two DL models, U-Net and the Siamese/Stereo matching network, based on \cite{bardozzo2022stasis}.
The experiments are performed on laparoscopic simulated stereo images \cite{bardozzo2022stasis}, with a focus on liver segmentation.

\subsubsection{Oncology}
Most of the explainable medical AI applications in image segmentation are in oncology.

\textbf{Liver}:
A DeepDream-inspired method is proposed in \cite{couteaux2019towards} for the segmentation of liver tumors in CT scans, specifically focusing on binary segmentation. The study seeks to understand how human-understandable features influence the segmentation output and defines the network’s sensitivity and robustness to these high-level features. High sensitivity indicates the importance of such features, while high network robustness shows its indifference to them. Radiomic features are also analyzed. The experiments are performed on the LiTS\footnote{https://competitions.codalab.org/competitions/17094} \cite{bilic2023liver} challenge dataset.
Semantic segmentation in liver CT images is further investigated in \cite{mohagheghi2022developing}, where the segmentation output is corrected based on XAI. This approach is categorized as a global surrogate and is model-agnostic. However, its primary purpose is not interpretability but rather the improvement in the initial segmentation by using additional boundary validation and patch segmentation models. The authors of \cite{dirks2022computer} investigate the segmentation of malignant melanoma lesions in 18-fluorodeoxyglucose (\textsuperscript{18}F-FDG) PET/CT modalities, focusing on metastasized tumors. The claim to interpretability is based on the visualization of the model's intermediate outcomes. The overall pipeline involves both segmentation and detection. Volumes of interest (VOI) are visualized for the liver as well as PET-positive regions classified as physiological uptake. This additional information is provided together with the final segmentation masks.

\textbf{Brain}:
An interpretable SUNet \cite{santamaria2020towards} architecture is proposed for the segmentation of brain tumors using The Cancer Imaging Archive (TCGA) dataset. Experimental results and statistical analysis indicate that symbolic sentences can be associated with clinically relevant information, including tissue type, object localization, morphology, tumor histology and genomics data. In \cite{saleem2021visual}, 3D visual explanations are investigated for brain tumor segmentation models, using the quantitative deletion curve metric to compare the results with Grad-CAM and Guided Backpropagation \cite{springenberg2014striving} techniques. In \cite{karri2022explainable}, a region-guided attention mechanism is used for the explainability of dermoscopic, multi-organ abdomen CT, and brain tumor MRI images. The experiments are performed on HAM10000 \cite{tschandl2018ham10000}, CHAOS 2019 \cite{kavur2021chaos}, and BraTS 2020 \cite{menze2014multimodal} datasets. Another architecture-based solution is proposed in \cite{schulze2022chimeric}, where the U-Net architecture is modified and applied to two MRI datasets: BraTS 2017~ \cite{menze2014multimodal} and OAI ZIB \cite{ambellan2019automated}, respectively focusing on brain tumors and human knees. In \cite{dasanayaka2022interpretable}, Grad-CAM results are compared to brain tumor segmentation results. The overall pipeline includes both classification and segmentation networks, where DenseNet is used for classification and Grad-CAM-based heatmaps are generated for different layers. However, Grad-CAM is not specifically tailored for segmentation but rather used as an explainable classification tool to evaluate segmentation results. In \cite{zeineldin2022explainability}, a NeuroXAI framework is introduced, combining seven backpropagation-based explainability techniques, each suitable for both explainable classification and segmentation. Gliomas and their subregions are investigated using 2D and 3D explainable sensitivity maps. A ProtoSeg method is proposed in \cite{he2023segmentation} for interpreting the features of U-Net, presenting a segmentation ability score based on the Dice coefficient between the feature segmentation map and the ground truth. Experiments are performed on five medical datasets, including BraTS for brain tumors, each focusing on different medical fields or affected organs.

\textbf{Pelvis}:
In \cite{dai2023explainable}, a  Generative Adversarial Segmentation Evolution (GASE) model is proposed for a multiclass 3D pelvis dataset \cite{dowling2015automatic}. The approach is based on adversarial training. Style-CAM is used to learn an explorable manifold. The interpretability part allows visualizing the manifold of learned features, which could  be used to explain the training process (i.e. what features are seen by the discriminator during training).

\textbf{Breast cancer}:
Oncological XAI applications for the segmentation of breast tumors are investigated in \cite{wang2023information}, \cite{karimzadeh2023post}, and \cite{farrag2023explainable}. In \cite{wang2023information}, a multitask network is proposed for both breast cancer classification and segmentation. Its interpretations are based on contribution score maps, which are generated by the information bottleneck. Three datasets are used, each focusing on a different imaging modality. In \cite{karimzadeh2023post}, SHAP explainability method is applied to the task of breast cancer detection and segmentation. The experiments are performed on BUSI \cite{al2020dataset}, BUSIS \cite{xian2018benchmark}, and HMSS \cite{geertsma2014ultrasoundcases} datasets. In \cite{farrag2023explainable}, explainability for mammogram tumor segmentation is investigated with the application of Grad-CAM and occlusion sensitivity, in both cases using Matlab implementations, and activation visualization. Their quantitative evaluation is based on image entropy, which gives additional information about the XAI method's complexity. Pixel-flipping techniques, which are directly related to deletion curves, are also employed. The experiments are performed on INbreast \cite{moreira2012inbreast} dataset of X-ray images.

\textbf{Other}: In \cite{zhang2023s}, Importance Activation Mapping (IAM) is employed as an explainable visualization technique in continual learning. The generated heatmap shows which regions in the input space are activated by model parameters with high-importance weights, associated with the model's memory. This approach is evaluated for the segmentation of prostate cancer. It also has applications in ophthalmology, specifically for segmenting the optic cup and disc. In \cite{dorrich2023explainable}, two CAM-based XAI techniques, Seg-Grad-CAM and High-Resolution CAM (HR-CAM), are applied to histopathological images of head and neck cancer. The explanations generated by both techniques appear to rely on the same features identified by professional pathologists. In \cite{cortacero2023evolutionary}, a solution based on Cartesian Genetic Programming is used to generate transparent and interpretable image processing pipelines. This method is applied to biomedical image processing, ranging from tissue histopathology to high-resolution microscopy images, and can be characterized as a few-shot learning approach. In \cite{kaur2022gradxcepunet}, a classification-based version of Grad-CAM is used to enhance a U-Net-based segmentation network. The experiments are performed on the 3D-IRCADb-01 \cite{christ2017automatic} dataset, comprised of 3D CT scans of venous phase CT patients. An Xception network generates 2D saliency maps for classification, which are then passed to the U-Net network together with the corresponding input images. This prior information enables more accurate segmentation. In \cite{pintelas2023xsc}, a framework for explainable classification and segmentation is presented. For segmentation, it relies on a feature hierarchy. The experiments are performed on the skin cancer dataset. The Factorizer architecture, introduced in \cite{ashtari2023factorizer}, is based on nonnegative matrix factorization (NMF) components, which are argued to be more semantically meaningful compared to CNNs and Transformers. The proposed approach is categorized under architecture-based interpretability methods. The models are implemented for brain tumor and ischemic stroke lesion segmentation datasets. In \cite{chatterjee2022torchesegeta}, a framework for explainable semantic segmentation is presented, extending several classification techniques to segmentation. These methods are also applied to 3D models. Infidelity and sensitivity metrics are used, and the experiments are performed on vessel segmentation in human brain images using Time-of-Flight Magnetic Resonance Angiogram. The experimental data \cite{mattern2018prospective} is not publicly available.  In \cite{kang2023learning}, a new interpretation method is proposed for multi-modal segmentation of tumors in PET and CT scans. It introduces a novel loss function to facilitate the feature fusion process. The experiments are performed on two datasets: a private non-small cell lung cancer (NSCLC) dataset and AutoPET \cite{gatidis2022whole}, a whole-body PET/CT dataset from MICCAI 2022 challenge.

\subsubsection{Ophthalmology}

XAI is also employed in the segmentation of ophthalmological images. 
Optic disc and cup segmentation is explored in the setting of continual learning \cite{zhang2023s}, where it is investigated in multi-site fundus datasets. Importance Activation Mapping is used to visualize the memorized content, facilitating an explanation of the model's memory. The focus is on reducing the model's forgetting. In \cite{he2023exploiting}, Seg-Grad-CAM is applied to ophthalmology for segmenting retinal layer boundaries. The study provides an entropy-based uncertainty visualization of segmentation probabilities. This offers more information about which retinal layers and regions exhibit higher uncertainty and allows for focusing on problematic areas. It is observed that higher uncertainty is associated with segmentation errors once it reaches a certain threshold. The experiments are performed on NR206\footnote{https://github.com/Medical-Image-Analysis/Retinal-layer-segmentation},

\subsubsection{Pancreatology}

In \cite{koker2021u}, an interpretable image segmentation approach is proposed for pancreas segmentation in CT scans. The method is also compared to Grad-CAM and occlusion sensitivity, demonstrating its superior inference time. This method identifies regions in the input images where noise can be applied without significantly affecting model performance. It relies on noisy image occlusion and can be classified as a perturbation-based technique. To directly parameterize the noise mask for each pixel without harming the model's performance, an additional small interpretability model is trained. Both interpretability and utility models are based on U-Net. Pixels that can be significantly perturbed without changing the model's performance are considered less important. Essentially, the proposed method involves training noise distributions. This approach allows training dynamic noise maps for individual images, differing from the typical static systematic occlusion. Experiments are performed on a pancreas dataset \cite{simpson1902large}. In \cite{okamoto2023generating}, a smoothing loss is introduced to guide interpretability learning. The authors observe that the explanations produced by U-Noise are less continuous. Assuming that important pixels are likely to be spatially close, the proposed smoothing objective considers the correlation between pixels during optimization. The resulting explanations are compared to those generated by Grad-CAM and U-Noise. Experiments are performed on a pancreas segmentation dataset \cite{antonelli2022medical} from the medical segmentation decathlon.

\subsubsection{Urology}

 In \cite{gao2023bayeseg}, a Bayesian approach is proposed to address the problem of interpreting domain-invariant features. The experiments are performed for prostate and cardiac segmentation tasks. The experiments are performed on T2 prostate MRI images from  NCI-ISBI 2013 \cite{bloch2015nci}, I2CVB \cite{lemaitre2015computer}, and PROMISE12 \cite{litjens2014evaluation}. For cardiac segmentation, MSCMR \cite{zhuang2018multivariate}, EMIDEC \cite{lalande2020emidec}, ACDC \cite{bernard2018deep}, MMWHS \cite{zhuang2016multi}, and CASDC 2013 \cite{kiricsli2013standardized} datasets are used.

\subsubsection{Anesthesiology}

In \cite{aguirre2023feet}, an interpretable approach is investigated for regional neuraxial analgesia monitoring.
The experiments focus on thermal foot images for patients who have received epidural anesthesia. The proposed method is based on Convolutional Random Fourier Features (CRFF) and layer-wise weighted CAM. The experiments are performed on the ThermalFeet\footnote{https://gcpds-image-segmentation.readthedocs.io/en/latest/notebooks/02-datasets.html} dataset of infrared images.

\subsection{Industry-related Applications}

Various industrial and industry-related activities require precise segmentation. These activities might range from precise manufacturing and processing \cite{gipivskis2023impact} to structural health monitoring in infrastructure, particularly in evaluating damage \cite{forest2023classification}. Here we discuss both industrial processes and indirectly related tasks, such as environmental monitoring and remote sensing, which can have potential in industrial applications in a more narrow sense. In this section, we divide industry-related explainable segmentation solutions into four categories: remote sensing, monitoring, scene understanding, and other more general applications.

\subsubsection{Remote Sensing}

One of the first applications of interpretable image segmentation is in remote sensing. In \cite{janik2019interpreting}, the U-Net model is applied for building detection. The proposed method works at the intersection of interpretability, representation learning, and interactive visualization, and is designed to explain U-Net's functionality. It employs Principal Component Analysis (PCA) on the activations in the bottleneck layer. PCA is the preferred method because it preserves the largest variance in the data. In the case of 3D visualizations, the first three components could be used. Following PCA, the new representations are clustered using k-means and DBSCAN algorithm. This approach allows for the visualization of learned latent representations for all samples through an Intersection over Union (IoU)-based heatmap, allowing users to identify qualitatively different regions. The experiments are performed on Inria Aerial Image Labeling (IAIL) \cite{maggiori2017can} dataset. This technique can be applied to detect and evaluate damages in industrial disasters or humanitarian crises, extending beyond mere infrastructure and product monitoring in industry. Another remote sensing application \cite{shreim2023trainable}, specifically focusing on high-resolution satellite images, employs a gradient-free Sobol method \cite{fel2021look} and a U-Noise model \cite{koker2021u}. The proposed method is also compared to Seg-Grad-CAM++ classification extension. 

\begin{table*}[htb]
  \newcommand{\z}{\phantom{0}}
  \newcommand{\za}{\phantom{,}}
  \newcolumntype{m}{>{\hsize=.7\hsize}X}

  \newcolumntype{b}{>{\hsize=.2\hsize}X}
  \newcolumntype{s}{>{\hsize=.2\hsize}X}
  \newcolumntype{L}{>{\raggedright}X}
  \centering
  \scriptsize
  \caption{Explainable Image Segmentation in Industry}
\begin{tabularx}{\textwidth}{L|LLsss} \hline
Category	&	Domain	&	Datasets & Metric &  	Year	&	Ref.	\\\hline
Remote sensing	&	Building detection	&	IAIL \cite{maggiori2017can} & $\vartriangleright$	&	2019	&	\cite{janik2019interpreting}	\\\hline
Scene understanding	&	Autonomous driving 	&	SYNTHIA \cite{zolfaghari2019temporal}, A2D2 \cite{geyer2020a2d2}	& $\vartriangleright$ &	2021	&	\cite{abukmeil2021towards}
\\\hline
Scene understanding	&	Pedestrian environments	&	PASCAL VOC 2012 \cite{everingham2010pascal}, ADE20K \cite{zhou2019semantic}, Cityscapes \cite{cordts2016cityscapes}	& NA* &	2021	&	\cite{zhang2021rethinking}	\\\hline
Scene understanding	&	Autonomous driving 	&	KITTI \cite{fritsch2013new}	& $\vartriangleright$ &	2022	&	\cite{mankodiya2022od}	\\\hline
Environmental monitoring	&	Flood detection	&	Worldfloods \cite{mateo2021towards}	& $\vartriangleright$ &	2022	&	\cite{zhang2022interpretable}	\\\hline
Scene understanding/ Biometrics	&	Driving scenes/Face recognition	&	BDD100k \cite{yu2020bdd100k}, CelebAMask-HQ \cite{lee2020maskgan}, CelebA \cite{liu2015deep}	& $\vartriangleright$ &	2022	&	\cite{jacob2022steex}	\\\hline
Monitoring/Scene understanding	&	Drones/Food processing	&	ICG drone dataset, private dataset	&	$\vartriangleright$ & 2023	&	\cite{gipivskis2023impact}	\\\hline
Monitoring/General applications	&	Food processing 	&	COCO \cite{lin2014microsoft}, private dataset	& $\vartriangleright$ &	2023	&	\cite{gipivskis2023ablation}	\\\hline
Biometrics	&	Facial emotions	&	Face recognition dataset \cite{turk1991eigenfaces} 	& $\vartriangleright$ &	2023	&	\cite{wang2023interpretable}	\\\hline
Monitoring	&	Cracks in infrastructure	&	CrackInfra \cite{liu2023transfer}	& $\vartriangleright$ &	2023	&	\cite{liu2023transfer}		\\\hline
General applications	&	Common objects	&	COCO \cite{lin2014microsoft}	& $\blacktriangleright$ &	2023	&	\cite{gipivskis2023occlusion}	\\\hline
Scene understanding/General applications	&	Street scenes/Common objects	&	Pascal VOC 2012 \cite{everingham2010pascal}, Cityscapes \cite{cordts2016cityscapes}	& $\vartriangleright$ &	2023	&	\cite{sacha2023protoseg}	\\\hline
Scene understanding	&	Driving scenes	&	BDD100k \cite{yu2020bdd100k}, BDD-OIA \cite{xu2020explainable}	& $\vartriangleright$ &	2023	&	\cite{zemni2023octet}	\\\hline
Scene understanding/General applications	&	Street scenes/Common objects	&	Cityscapes \cite{cordts2016cityscapes}, Pascal VOC \cite{everingham2010pascal}, COCO \cite{lin2014microsoft}	& $\blacktriangleright$ &	2023	&	\cite{dreyer2023revealing}	\\\hline
General applications	&	Common objects	&	COCO \cite{lin2014microsoft} 	& $\vartriangleright$ &	2023	&	\cite{hasany2023seg}	\\\hline
General applications	&	Common objects	&	Pascal VOC \cite{everingham2010pascal}	& $\blacktriangleright$ &	2023	&	\cite{cheng2023tax}	\\\hline
Scene understanding/Remote sensing	&	Street scenes/Building detection	&	Cityscapes \cite{cordts2016cityscapes}, WHU \cite{ji2018fully}	& $\blacktriangleright$ &	2023	&	\cite{shreim2023trainable}	\\\hline

\multicolumn{6}{l}{*The application focuses on introducing explainability to segmentation evaluation, rather than evaluating explainability techniques.}\\
\multicolumn{6}{l}{$\vartriangleright$:  Qualitative XAI evaluation}\\
\multicolumn{6}{l}{$\blacktriangleright$:  Quantitative XAI evaluation}\\
   \end{tabularx} 
\label{tab:industry}
\end{table*} 

\subsubsection{Monitoring}

Here we review relevant papers that offer explainable segmentation-based monitoring in proximate environments. In \cite{gipivskis2023impact}, simple gradient \cite{simonyan2013deep} saliency maps and SmoothGrad-based \cite{smilkov2017smoothgrad} saliencies are implemented for semantic segmentation models to investigate the adversarial attack setting. The experiments are performed on two industry-related cyber-physical system datasets. A private dataset from CTI FoodTech, a manufacturer of fruit-pitting machines, is used. In \cite{gipivskis2023ablation}, the same private dataset is used for experiments with gradient-free XAI technique, based on the perturbations of intermediate activation maps. 

In \cite{liu2023transfer}, the focus is on crack segmentation in critical infrastructures, such as tunnels and pavements. The U-Net model is used together with Grad-CAM, which is applied at the bottleneck, as in \cite{vinogradova2020towards}. They investigate both simple and complex crack patterns as well as different backgrounds. Two other papers \cite{seibold2022explanations}, \cite{forest2023classification} also investigate the segmentation of different crack types. However, the proposed XAI techniques are implemented in classification models, and used for weakly supervised segmentation. These techniques are discussed in the subsequent section. In \cite{wang2023interpretable}, an interpretable Bayesian network is used for facial micro-expression recognition. The authors prefer these networks for segmentation over DL models, primarily because of their superior causal interpretability when dealing with uncertain information. This can make them better interpretability candidates when uncertain causal inference is involved. The experiments are performed on the database \cite{turk1991eigenfaces} of face images.

\subsubsection{Scene Understanding}

Scene understanding is an important area in applications for autonomous vehicles, monitoring of pedestrians and ambient objects, and surveillance. Precise real-time segmentation of road signs and obstructions is of particular importance.
Explainable segmentation can be seen as part of explainable autonomous driving systems  \cite{abukmeil2021towards}, which investigate events, environments, and engine operations. An explainable variational autoencoder (VAE) model is proposed in \cite{abukmeil2021towards}, focusing on neuron activations with the use of attention mapping. For the experiments, the SYNTHIA \cite{zolfaghari2019temporal} and A2D2 \cite{geyer2020a2d2} datasets are used. The results are analyzed both qualitatively and quantitatively, using the average area under the
receiver operator characteristic curve (AUC-ROC) index. In \cite{mankodiya2022od}, XAI techniques are employed to investigate pixel-wise road detection for autonomous vehicles. The experiments are performed on different segmentation models, using the KITTI \cite{fritsch2013new} road dataset. The problem is formulated as a binary segmentation task, where the classes are limited to the road and its surroundings. Grad-CAM and saliency maps are used to generate explanations.  Unmanned aerial vehicles can also fall under the category of autonomous driving systems. In \cite{gipivskis2023impact},  gradient-based XAI techniques are applied to semantic drone dataset\footnote{http://dronedataset.icg.tugraz.at/} from Graz University of Technology.

Automated semantic understanding of pedestrian environments is investigated in \cite{zhang2021rethinking}. Here the focus is not on a particular XAI technique, but on introducing some level of explainability to segmentation evaluation. The paper argues that popular pixel-wise segmentation metrics, such as IoU or Dice coefficient, do not sufficiently take into account region-based over- and under-segmentation. Here over-segmentation refers to those cases where the relevant ground-truth region is segmented into a lower number of regions than the predicted mask. For instance, where there is only one bus in the segmented ground-truth, but the model segments it into three disjoint segments. In the case of under-segmentation, the opposite is true. Pixel-wise metrics do not accurately represent these differences in disjoint and joint regions as long as a large enough number of similar pixels is segmented in both the ground-truth image, and the corresponding prediction. The use of region-wise measures is proposed as a better way to explain the source of error in segmentation. The experiments are performed on PASCAL VOC 2012 \cite{everingham2010pascal}, ADE20K \cite{zhou2019semantic}, and Cityscapes \cite{cordts2016cityscapes}. In \cite{di2023sediment}, the focus is on automatic semantic segmentation for sediment core analysis. To interpret the results, higher segmentation error regions and model prediction confidence are visualized. Here, the model confidence is defined as prediction probability, and the model error calculation is based on the normalized categorical cross-entropy.

The authors of \cite{dreyer2023revealing} propose the Concept Relevance Propagation-based approach L-CRP as an extension of CRP \cite{achtibat2022towards}. By utilizing concept-based explanations, the study seeks to gain insights into both global and local aspects of explainability. The proposed approach seeks to understand the contribution of latent concepts to particular detections by identifying them, finding them in the input space, and evaluating their effect on relevance. Context scores are computed for different concepts. The experiments are performed on Cityscapes \cite{cordts2016cityscapes}, Pascal VOC \cite{everingham2010pascal}, and COCO \cite{lin2014microsoft} datasets.

\subsubsection{General Applications}

Some of the XAI-related experiments focus on more general datasets, typically used in evaluating the performance of segmentation models. COCO \cite{lin2014microsoft} dataset has been used as a benchmark in \cite{gipivskis2023occlusion} and \cite{gipivskis2023ablation}. The dataset is composed of 21 classes of everyday objects, including several types of vehicles. Both \cite{gipivskis2023occlusion} and \cite{gipivskis2023ablation} apply perturbation-based gradient-free methods. Input perturbations are used in \cite{gipivskis2023occlusion}, while feature map perturbations in pre-selected intermediate layers are used in \cite{gipivskis2023ablation}.

 Tendency-and-Assignment Explainer (TAX) framework is introduced in \cite{cheng2023tax}. It seeks to explain: 1) what contributes to the segmentation output; and 2) what contributes to us thinking so (i.e. the why question). For this, a multi-annotator scenario is considered. The learned annotator-dependent prototype bank indicates the segmentation tendency, with a particular focus on uncertain regions. The experimental results on the Pascal VOC \cite{everingham2010pascal} dataset demonstrate that TAX predicts oversegmentation consistent with the annotator tendencies. 

\subsection{XAI Applications in Self-Supervised and Weakly Supervised Segmentation}

Manual image labeling is an expensive operation, especially when pixel-wise labeling is involved. It requires significant time and financial resources, and depending on the dataset being annotated, may also require particularly narrow expertise. With this in mind, it has been suggested that XAI techniques could be employed for automated labeling, which could also help reduce some forms of annotation bias.

In \cite{yu2023ex}, a new explainable transformer architecture is proposed for model-inherent interpretability. The proposed model, a Siamese network, is investigated for weakly supervised segmentation. For enhanced interpretability, model representations are regularized using an attribute-guided loss function. Higher-layer attention maps are fused and used alongside attribute features. Qualitative segmentation results are compared with the SIPE technique. However, the model's limitation is its inability to incorporate attribute-level ground truth labels. Another application for weakly supervised segmentation employs LRP-based classification explanations \cite{seibold2022explanations}. These explanations are used to generate pixel-wise binary segmentations, which are then thresholded. The experiments are conducted on two datasets: one for cracks in sewer pipes and another for cracks in magnetic tiles \cite{huang2020surface}.

In \cite{forest2023classification}, surface crack detection and growth monitoring are investigated as part of structural health evaluation in infrastructure. Although no specific technique for explainable segmentation is proposed, explainable classification is used for weakly supervised segmentation, allowing for the quantification of crack severity. Six post-hoc techniques are implemented: InputXGradient, LRP, Integrated Gradients, DeepLift, DeepLiftShap, and GradientShap. Additionally, B-cos networks and Neural Network Explainer are employed. In \cite{bedmutha2021using}, GradCAM is used for semantic segmentation. An additional classification model is used with masked inputs, based on the given class. The classifier is trained on all the masked images across all classes. Explainable classification can also be used to enhance the data efficiency of segmentation models. For instance, in \cite{wu2024dea}, Grad-CAM is employed to extract data-efficient features from the classification model, which are then used for segmentation. The results indicate that this approach generalizes across different segmentation methods.

\section{Discussion}
\subsection{Open Issues}
Plenty of unresolved challenges remain in explainable semantic segmentation, most of which are also applicable to image classification tasks. Below is a non-exhaustive list of these challenges:
\renewcommand{\thesubsubsection}{\alph{subsubsection}}

\begin{itemize}
\item \textbf{Evaluation metrics for XAI}
Most of the literature on XAI in image classification focuses on introducing new explainability techniques and their modifications, rather than proposing new evaluative frameworks or benchmark datasets. This tendency is even more visible in explainable semantic segmentation. Currently, there are no papers dedicated solely to evaluating XAI results in image segmentation. The investigation of XAI metrics remains limited to the experimental results sections, and only in those few cases where quantitative evaluation is used. There is no consensus on which evaluation metrics are most crucial for capturing the key aspects of explainability, largely due to the difficulty in formalizing explainability-related concepts. A better theoretical understanding of the problem should inform the creation of evaluative XAI metrics and benchmarks. Such foundations would likely result in more efficient explainable segmentation methods that are better adapted to the problem at hand.

\item \textbf{Safety and robustness of XAI methods}
With the rapid deployment of DL models in medical, military, and industrial settings, XAI techniques are set to play an even more important role. Their primary use is driven by the need to determine if the model is reliable and trustworthy. However, similar questions can also be raised about the XAI techniques themselves. It is important to investigate their vulnerabilities and loopholes. Both deployers and end-users need to know whether they are secure against intentional attacks directed at XAI techniques or the model. Even if there is no direct threat, the robustness of each specific XAI method needs to be investigated on a case-by-case basis.

Just like classification models, semantic segmentation models are susceptible to adversarial attacks. Different attack methods have been proposed \cite{fischer2017adversarial}, \cite{cisse2017houdini}, \cite{xie2017adversarial}. When discussing adversarial attacks, it is common to focus on the model's output as the primary target. However, it is also possible to attack the output's explanation saliency while leaving both the input and the output perceptibly unchanged. Such attacks have been introduced and investigated in the context of image classification \cite{dombrowski2019explanations}. It has also been demonstrated that these second-level attacks can be extended to image segmentation \cite{gipivskis2023impact}. More research is needed to find the best ways to combat them, especially since new adversarial attacks are constantly being developed, and comprehensive safety guarantees are challenging to ensure. Systematic investigations need to be undertaken for both white-box attacks, where the attacked model is known to the attacker, and black-box attacks, where it is unknown. Similar investigations into the robustness of interpretable segmentation could contribute to the overall security of AI systems.

Adversarial examples are typically not part of the training and testing datasets. This omission can lead to vulnerabilities in deployed models. Another critical issue is the presence of biases. When the most salient regions of the explanation map fall outside the boundaries of the object of interest, this might signal not just a misguided prediction but also the potential presence of adversarial influences \cite{hoyer2019grid}. Natural adversarial examples \cite{hendrycks2021natural} and their influence on XAI in segmentation could be investigated as well.

\item \textbf{XAI for video segmentation}
As semantic scene segmentation is not limited to 2D images, new interpretability techniques could be investigated for video data, where temporal semantic segmentation is carried out. Video object segmentation requires significantly more computational resources. To our knowledge, there are currently no studies investigating explainable image segmentation in a dynamic setting. The nature of dynamic scenes could introduce novel challenges not previously encountered in 2D segmentation contexts. For instance, one would need to add an additional temporal explanation axis to account for differences in interpretability maps across video frames. This task could be further extended to real-time semantic segmentation by focusing on ways to reduce the latency of the generated explanations.
\end{itemize}

\subsection{Future Directions}
Given that most literature primarily focuses on qualitative metrics, we aim to highlight the need for a well-defined benchmark and evaluation strategy for XAI methods in image segmentation. To our knowledge, there are currently no studies on evaluation metrics or benchmarks specifically for XAI methods in this area. Moreover, research focusing on the formal aspects of quantitative metrics in XAI is limited. We see mechanistic interpretability \cite{cammarata2021curve} as a promising research area. This approach seeks to reverse-engineer how models function. As far as we are aware, there have been no significant contributions in formal explainability \cite{marques2023logic} or argumentative XAI \cite{vcyras2021argumentative} within the context of image segmentation. Additionally, there has not been much research \cite{karim2023understanding} into the interpretability of transformers for segmentation, especially compared to convolutional networks. 

\begin{itemize}

\item \textbf{Failure Modes}
This area is related to evaluation metrics. However, it covers problematic areas that could not be identified by the commonly used metrics. Specifically, XAI could be used to identify and mitigate bias in segmentation models. A systematic analysis of failure cases and potential failure modes could better determine the scope of applicability for XAI methods. Several studies \cite{adebayo2018sanity} have critically evaluated different groups of explainability techniques in classification. However, a similar investigation has not yet been conducted in image segmentation. 

\item \textbf{Neural architecture search}
Neural architecture search (NAS) explores automating neural architecture designs. XAI techniques can be applied in NAS in at least two distinct ways. First, existing XAI methods can be incorporated into NAS algorithms to improve their performance. For example, in \cite{zhang2023cam}, an explainable CAM technique is integrated with the NAS algorithm to avoid fully training submodels. Second, NAS algorithms can include interpretability aspects as one of the metrics to be optimized in multi-objective optimization. In \cite{carmichael2021learning}, a surrogate interpretability metric has been used for multi-objective optimization in image classification. However, currently, no similar approaches exist for semantic segmentation tasks.

\item \textbf{Continual Learning}
Continual learning (CL) refers to the research area that investigates techniques allowing models to learn new tasks without forgetting the previously learned ones. This strong tendency for DL models to forget previously learned information upon acquiring new knowledge is commonly described as catastrophic forgetting. More efficient solutions to CL problems would allow the models to be used more resourcefully, without retraining them from scratch when new data arrives. The intersection of XAI and CL presents an interesting area for investigation. XAI methods can be employed in CL to: 1) improve the model's performance; 2) better understand and explain the model's predictions; and 3) investigate the phenomenon of catastrophic forgetting. The exploration of XAI and CL could also lead to improved model understanding when either a shift in data distribution or concept drift occurs.

\end{itemize}

\section{Conclusion}

 This survey presents a comprehensive view of the field of XAI in image segmentation. Our goal has been twofold: first, to provide an up-to-date literature review of various types of interpretability methods applied in semantic segmentation; and second, to clarify conceptual misunderstandings by proposing a method-centered taxonomy for image segmentation and general frameworks for different types of interpretability techniques. To these ends, we have categorized the methods into five major subgroups: prototype-based, gradient-based, perturbation-based, counterfactual methods, and architecture-based techniques. Based on the surveyed literature on explainable image segmentation, it is evident that most of the methods focus on local explanations and rely on qualitative evaluation. We hope this work can benefit computer vision researchers by presenting the landscape of XAI in image segmentation, delineating clearer boundaries between existing methods, and informing the development of new interpretability techniques.

\bibliographystyle{IEEEtran}
\bibliography{manuscript}

\end{document}